\documentclass[11pt]{article}
\pdfmapfile{+cm.map}
\pdfmapfile{+cmextra.map}
\pdfmapfile{+latxfont.map}
\pdfmapfile{+symbols.map}

\usepackage[margin=1in]{geometry}
\usepackage{amsmath,amssymb}
\usepackage{booktabs}
\usepackage{graphicx}
\graphicspath{{figures/}}
\usepackage{xcolor}
\usepackage{hyperref}
\usepackage{url}
\usepackage{enumitem}
\usepackage{caption}
\usepackage{subcaption}
\usepackage{multirow}

\hypersetup{
  colorlinks=true,
  linkcolor=blue!55!black,
  citecolor=blue!55!black,
  urlcolor=blue!55!black,
  pdftitle={Every Expert Counts: ExactMoE for Memory-Efficient W4A16 Inference},
  pdfauthor={Amjad Saab},
  pdfsubject={Memory-efficient mixture-of-experts inference},
  pdfkeywords={mixture of experts, quantization, W4A16, MARLIN, inference}
}
\setlist{nosep,leftmargin=*}
\newcommand{\method}{\textsc{ExactMoE}}

\title{\textbf{Every Expert Counts: ExactMoE for Memory-Efficient W4A16 Inference}}
\author{
Amjad Saab\\
Appendture\\
Dubai, United Arab Emirates\\
\texttt{amjad.saab@appendture.com}
}
\date{\small August 4, 2026}

\begin{document}
\maketitle

\begin{abstract}
Sparse mixture-of-experts (MoE) language models reduce arithmetic by activating only a small subset of experts per token, yet deployment still requires storing and moving the full expert bank. We present \method{}, an inference design that applies symmetric group-128 four-bit weight quantization only to routed experts, stores those experts in kernel-native MARLIN form in pinned host memory, and executes all selected experts through a configurable GPU-resident slot cache and fused grouped MoE kernels. The router, attention, embeddings, normalization layers, and language-model head remain in BF16. ``Exact'' refers to complete expert availability and an unchanged top-$k$ routing procedure: no expert is pruned, substituted, or forced to execute on the CPU. It does not imply numerical identity with the BF16 model.

On OLMoE-1B-7B-0924-Instruct, evaluated on a single NVIDIA L4, a 16-slot configuration reduces peak reserved GPU memory from 14.168 to 1.836~GiB (87.04\%) while retaining 81.85\% of BF16 decode throughput. A fully resident 64-slot configuration reaches 31.923 tokens/s versus 21.662 tokens/s for BF16 while reserving 4.061~GiB. Across 12,450 zero-shot multiple-choice questions, \method{} obtains 70.3534\% normalized accuracy versus 70.8996\% for BF16, retaining 99.23\% of the baseline accuracy. In a matched 16-token ablation, fused grouped execution is $1.97\times$ as fast as a sequential W4 reference. These results identify a practical memory--transfer--throughput frontier for complete-expert MoE inference.
\end{abstract}

\section{Introduction}

Sparse MoE models increase parameter capacity without activating every parameter for every token \cite{fedus2022switch}. This conditional computation makes training and arithmetic efficient, but it does not eliminate the deployment cost of the inactive experts. A runtime must either keep a large expert bank on the accelerator, move experts through a slower memory tier, or reduce the expert representation. Each choice introduces a different bottleneck: GPU capacity, host-to-device traffic, or quantization error.

Existing work attacks parts of this problem. Expert-only quantization exploits the empirical robustness of MoE experts \cite{kim2023moqe}; mixed-precision methods allocate different bit widths or placements to experts \cite{imani2024mixture,tang2024hobbit}; and offloading systems use activation history and caching to reduce transfers \cite{xue2024moeinfinity}. General-purpose weight-only kernels such as MARLIN provide fast W4A16 matrix multiplication \cite{frantar2024marlin}. The remaining systems question is whether these ingredients can be combined while preserving the model's full expert set and avoiding slow per-expert execution or transient FP16 materialization.

We study that question with \method{}. The routed expert bank is quantized once into the exact packed format consumed by the GPU kernel. The same packed representation is stored in pinned host memory and in a configurable GPU cache, so a cache miss copies an expert without dequantizing or repacking it. At each sparse layer, active global expert identifiers are resolved in one batched transfer. If the active set exceeds the cache capacity, it is partitioned into resident waves. Each wave is executed by a fused multi-expert MARLIN kernel using an explicit global-to-slot map, and the wave outputs are accumulated. All selected experts therefore contribute to the result.

Our contribution is empirical and systems-oriented rather than theoretical. Expert-only quantization, LRU-style caching, and MARLIN are not individually new. The contribution is their integration into a complete-expert runtime with a common kernel-native host/GPU representation, grouped fused execution, and a measured capacity frontier. Specifically:

\begin{itemize}
    \item We define a complete-expert W4A16 execution contract that leaves the router and dense model components in BF16 while quantizing only routed expert projections.
    \item We implement kernel-native offline packing, pinned host storage, configurable slot residency, batched expert-ID resolution, and wave-based fused execution without FP16 expert materialization.
    \item We provide a controlled OLMoE study comparing BF16 and \method{} on the same checkpoint and hardware, including quality, latency, throughput, memory, transfer volume, cache behavior, and paired statistics.
    \item We separate three notions that are often conflated: complete expert availability, routing faithfulness for the quantized hidden state, and numerical equivalence to BF16. \method{} guarantees the first two, not the third.
\end{itemize}

\begin{figure}[t]
    \centering
    \includegraphics[width=\linewidth]{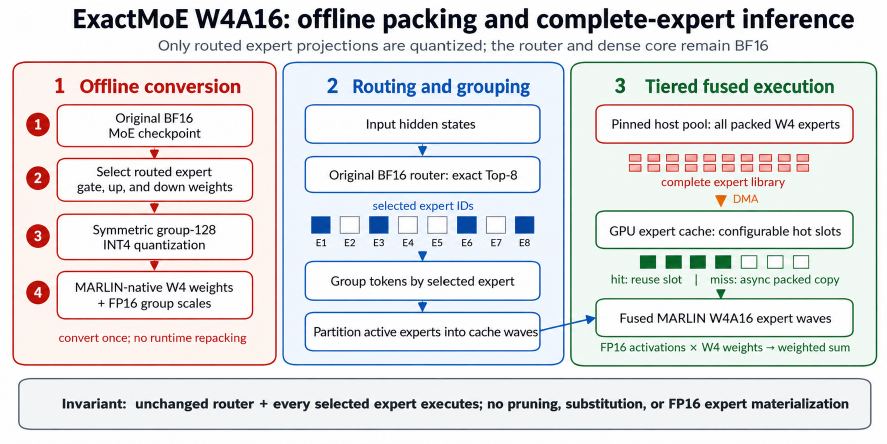}
    \caption{Overview of \method{}. Routed expert projections are quantized and packed once into the representation consumed by MARLIN. During inference, the original BF16 router selects experts, tokens are grouped by selected expert, and active experts execute in cache-sized fused waves. The pinned host pool retains every packed expert; GPU cache misses transfer packed weights and scales without FP16 expert materialization.}
    \label{fig:method-overview}
\end{figure}

Figure~\ref{fig:method-overview} summarizes the conversion and inference paths. The diagram is intentionally separated into offline representation building, router-preserving token grouping, and tiered fused execution so that the method's invariants are distinct from its empirical performance claims.

\section{Related Work}

\paragraph{Sparse mixture-of-experts.}
Switch Transformers established a widely used sparse-routing design \cite{fedus2022switch}. OLMoE provides fully open weights, data, training code, and logs for a 7B-parameter model with approximately 1B active parameters \cite{muennighoff2024olmoe}. Qwen1.5-MoE and later Qwen MoE families provide additional architectures with shared and routed experts \cite{bai2023qwen,qwen2024moe,qwen2026qwen36}. These models differ in checkpoint layout, expert geometry, routing policy, attention architecture, and auxiliary components, making portable expert-only inference nontrivial.

\paragraph{MoE quantization.}
MoQE showed that expert weights can tolerate low-bit quantization and that expert-only quantization can reduce model size and latency \cite{kim2023moqe}. Mixture of Experts with Mixture of Precisions studies precision allocation as a quality-of-service control \cite{imani2024mixture}. More recent methods optimize precision by expert sensitivity or activation frequency, add low-rank compensation, or adapt expert precision at runtime \cite{duanmu2025mxmoe,huang2025milo,chu2025dynaexq,yang2026pagedweight}. These works make clear that expert quantization is not itself new. \method{} instead studies a fixed uniform W4 representation that is stored in the same kernel-native form in host and GPU memory while keeping every routed expert available.

\paragraph{Expert offloading and caching.}
MoE-Infinity exploits request-level activation locality for expert caching and prefetching \cite{xue2024moeinfinity}. SwapMoE maintains a tunable set of virtual experts, Fiddler orchestrates CPU and GPU computation to reduce transfers, HOBBIT uses mixed-precision experts to reduce miss latency, and MoE-Lightning pipelines CPU, GPU, and I/O work for throughput-oriented batch inference \cite{kong2024swapmoe,kamahori2025fiddler,tang2024hobbit,cao2025moelightning}. Our work shares the tiered-residency objective, but uses one W4 representation in every tier and executes every selected expert on the GPU. We do not claim superiority over these systems because they were not reproduced under the same hardware and workload protocol.

\paragraph{Weight-only kernels and serving runtimes.}
GPTQ and AWQ are influential post-training weight quantization methods for dense transformers \cite{frantar2022gptq,lin2024awq}. MARLIN provides mixed-precision W4A16 kernels designed to retain speedups over a range of batch sizes \cite{frantar2024marlin}. vLLM provides high-throughput serving and memory management through PagedAttention \cite{kwon2023vllm}. \method{} uses MARLIN-format packed weights and vLLM's fused MARLIN MoE primitive, but contributes the expert-cache management and complete-expert grouped execution around that primitive.

\section{Method}

\subsection{Scope and terminology}

Consider sparse layer $\ell$ with $E$ routed experts and top-$k$ routing. Given hidden state $h_t^{(\ell)}$, the router computes logits
\begin{equation}
    z_t^{(\ell)} = W_{r}^{(\ell)}h_t^{(\ell)}, \qquad
    \mathcal{S}_t^{(\ell)} = \operatorname{TopK}\!\left(z_t^{(\ell)}, k\right),
\end{equation}
with router-derived mixture weights $p_{t,e}^{(\ell)}$. The sparse block output is
\begin{equation}
    y_t^{(\ell)} = \sum_{e\in\mathcal{S}_t^{(\ell)}} p_{t,e}^{(\ell)}
    W_{d,e}^{(\ell)}\left[
    \phi\!\left(W_{g,e}^{(\ell)}h_t^{(\ell)}\right)
    \odot W_{u,e}^{(\ell)}h_t^{(\ell)}\right].
    \label{eq:moe}
\end{equation}

\method{} preserves $W_r$, the router's top-$k$ rule, and every expert index $e\in\{1,\ldots,E\}$. It replaces only $W_g$, $W_u$, and $W_d$ with groupwise INT4 representations whose dequantized values define the corresponding mathematical operations; the implementation does not materialize full FP16 expert matrices. Quantization changes hidden states, so later routing decisions can differ from a BF16 rollout. Our use of \emph{routing-faithful} means that the original router is evaluated on the current quantized-model hidden state and that all of its selected experts execute.

\subsection{Expert-only symmetric W4 quantization}

For each expert matrix $W\in\mathbb{R}^{N\times K}$, we partition the input dimension into groups $G$ of 128 weights. For row $i$ and group $g$, we compute
\begin{equation}
    s_{i,g}=\frac{\max_{j\in g}|W_{i,j}|}{7},\qquad
    q_{i,j}=\operatorname{clip}\left(\operatorname{round}\left(\frac{W_{i,j}}{s_{i,g}}\right),-8,7\right),
    \label{eq:quant}
\end{equation}
and use $\widehat{W}_{i,j}=s_{i,g}q_{i,j}$ as the mathematical reference. We then pack $q$ and FP16 scales into MARLIN-native tensors. Gate and up projections are concatenated in the ordering expected by the fused kernel; down projections are packed separately. Quantization and packing happen offline once per checkpoint conversion.
For an all-zero group, we define $s_{i,g}=0$ and $q_{i,j}=0$ directly, avoiding the division in Equation~\ref{eq:quant}.

The routed expert pool requires approximately
\begin{equation}
    M_{\mathrm{W4}} \approx P_{\mathrm{expert}}\left(\frac{4}{8}+\frac{16}{128\cdot 8}\right)\ \text{bytes},
\end{equation}
where $P_{\mathrm{expert}}$ is the routed-expert parameter count. For the 6.44B routed-expert parameters in OLMoE, the measured packed host pool is 3.09~GiB.

\subsection{Tiered expert residency}

Each sparse layer owns a cache with $C$ expert slots, $k\le C\le E$. A slot stores packed gate/up weights, packed down weights, and their FP16 scales. The host pool stores the same objects in pinned memory. Consequently, a miss performs a direct asynchronous copy of the final kernel representation; it never materializes an FP16 expert.

The cache maintains a global-expert-to-slot map and a recency timestamp. For a set of active experts, resident entries become hits, missing entries replace the least-recently-used slots not protected by the current wave, and CUDA events prevent a slot from being overwritten before its preceding computation finishes. When $C=E$, all experts are preloaded and expert-ID transfers and cache replacement are disabled.

\subsection{Grouped complete-expert execution}

One decode token can activate at most $k$ experts per layer, but prefill or batched generation can activate more than $C$ distinct experts across tokens. We therefore partition the layer's active expert set into waves of at most $C$ experts. For each wave, \method{}:

\begin{enumerate}
    \item resolves or loads every wave expert into a slot;
    \item constructs an expert map from global IDs to resident slot IDs and maps every other expert to an inactive sentinel;
    \item calls a fused MARLIN MoE kernel for the complete token matrix;
    \item accumulates the wave output into the layer result; and
    \item records completion events before any wave slot may be reused.
\end{enumerate}

The production path therefore performs two grouped expert GEMMs per wave---one fused gate/up projection and one down projection---rather than a Python or kernel call per active expert. Router weights are retained across waves and are not renormalized within a wave. If the active set spans several waves, Equation~\ref{eq:moe} is recovered by summing their disjoint contributions.

\subsection{Correctness gates}

We distinguish three checks. First, a kernel test compares MARLIN output with the materialized dequantized W4 reference; it tests packing and kernel correctness, not BF16 quantization error. Second, a fused-versus-sequential W4 ablation checks exact greedy tokens and final-logit tolerance under a matched cache state. Third, end-to-end BF16 comparisons measure the actual quality effect of expert quantization. Fresh-process export/reload parity is recorded separately as a deployment check in the artifact archive.

\section{Experimental Setup}

\subsection{Model and software}

The controlled study uses
\path{allenai/OLMoE-1B-7B-0924-Instruct} at revision
\texttt{7f1c97f440f06ce3\allowbreak
6705e4f2b843edb5\allowbreak
925f4498} \cite{muennighoff2024olmoe}.
The model contains 16 sparse layers with 64 experts per layer, top-$8$
routing, a hidden size of 2048, and an expert intermediate size of 1024.
Its routed experts contain 6.44 billion parameters. The BF16 baseline and
\method{} use the same checkpoint, tokenizer, SDPA attention backend,
prompt set, and greedy decoding policy.

Experiments were conducted on one NVIDIA L4 GPU with 22.0~GiB of usable
GPU memory and 53~GiB of host memory. The recorded software environment
included PyTorch 2.11.0+cu130, CUDA 13.0, Transformers 5.14.1, and
vLLM 0.26.0. The packing implementation was aligned with
\path{IST-DASLab/marlin} commit \texttt{1f25790}, while grouped expert
execution used vLLM's fused MARLIN MoE primitive. All OLMoE results reported in this paper were produced
during the same controlled experiment; measurements from earlier experiments
on other model architectures are not included in the OLMoE tables.

\subsection{Configurations and metrics}

We evaluate cache capacities $C\in\{8,16,32,64\}$. Capacity 64 is fully resident. The primary offloaded comparison uses $C=16$. Runtime evaluation uses 10 fixed prompts, three repetitions, and 128 generated tokens per request. We report decode throughput, time to first token (TTFT), peak reserved GPU memory, cache events, and transferred bytes. Peak reserved memory is the maximum memory held by the PyTorch CUDA allocator; it is not model-file size or total board memory. ``Persistent warm'' means process and kernels remain initialized across requests; cache contents persist according to the configured policy. Model loading and offline expert packing are excluded from request timing.

The batch sweep uses fixed batches $\{1,2,4,8\}$ and 32 generated tokens per request. It measures ordinary batched generation, not a continuous-batching server. Each batch point is a single sweep measurement and is treated as descriptive.

\subsection{Quality evaluation}
\label{sec:quality-eval}

We evaluate all 570 ARC-Easy validation questions \cite{clark2018arc}, all 1,838 PIQA validation questions \cite{bisk2020piqa}, and all 10,042 HellaSwag validation questions \cite{zellers2019hellaswag}. For each answer choice, we compute continuation log-likelihood under the model's chat template. Normalized accuracy selects the highest mean token log-likelihood; raw accuracy selects the highest summed log-likelihood. The aggregate contains 12,450 paired questions.

We additionally report teacher-forced likelihood on 120 WikiText-2 test passages \cite{merity2016wikitext} and 120 deterministic samples from Dolly-15k \cite{conover2023dolly}. These are immutable custom manifests and are not claimed to reproduce lm-evaluation-harness scores. We use 10,000 paired bootstrap resamples for accuracy-difference intervals and an exact two-sided McNemar test on paired predictions.

\subsection{Evaluation scope}

The controlled comparison isolates the effect of the W4A16 expert representation and cache capacity on one checkpoint and one GPU. Prompt repetitions share a persistent process and therefore estimate within-process variability rather than fresh-process or cross-machine variance. The batch sweep contains one measurement at each batch size and does not use a request scheduler. We compare against the original BF16 checkpoint and an internal sequential W4 reference; no external MoE quantization or offloading system was reproduced on the same hardware. The results therefore establish the measured quality--memory--throughput trade-off of this implementation, not general superiority over existing serving systems.

\section{Results}

\subsection{Memory--throughput frontier}

\begin{table}[t]
\centering
\caption{Persistent-warm OLMoE runtime medians. H2D is median host-to-device traffic per 128-token request. Full residency preloads all experts before timing.}
\label{tab:capacity}
\small
\begin{tabular}{lrrrrrr}
\toprule
Method & Slots & Decode & TTFT & Reserved & Hit rate & H2D \\
 & & (tok/s) & (ms) & (GiB) & (\%) & (GiB) \\
\midrule
BF16 & -- & 21.66 & 79.03 & 14.17 & -- & 0.00 \\
\method{} & 8  & 15.76 & 340.77 & 1.48 & 38.8 & 30.04 \\
\method{} & 16 & 17.73 & 290.42 & 1.84 & 56.3 & 21.44 \\
\method{} & 32 & 21.66 & 210.74 & 2.58 & 81.4 & 9.13 \\
\method{} & 64 & \textbf{31.92} & \textbf{33.92} & 4.06 & resident & 0.00 \\
\bottomrule
\end{tabular}
\end{table}

\begin{figure}[t]
    \centering
    \includegraphics[width=\linewidth]{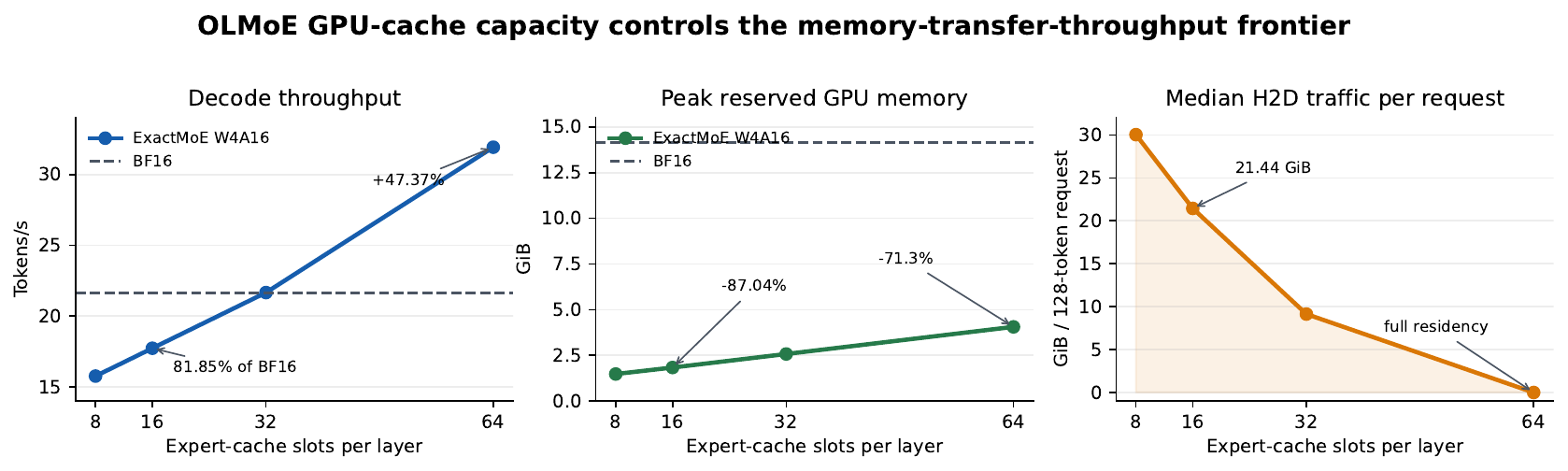}
    \caption{Measured OLMoE capacity frontier from the archived Colab run. Larger caches reduce expert transfers and improve decode throughput at the cost of additional GPU memory. Dashed lines show the matched BF16 reference. Capacity 64 is fully resident; its zero hit rate in raw counters means cache lookup accounting is bypassed, not that experts are absent.}
    \label{fig:frontier}
\end{figure}

Table~\ref{tab:capacity} and Figure~\ref{fig:frontier} show a monotonic capacity frontier. At $C=16$, peak reserved GPU memory falls from 14.168 to 1.836~GiB, an 87.04\% reduction, while decode throughput is 17.731 tokens/s, or 81.85\% of BF16. The cost is 21.44~GiB of expert traffic per measured request and a 290~ms median TTFT. Increasing capacity to 32 raises the hit rate to 81.4\%, nearly matches BF16 decode throughput, and still reduces reserved memory by 81.8\%. At full residency, the packed expert bank uses 4.061~GiB reserved memory and reaches 31.923 tokens/s, 47.4\% above BF16. Full residency should be interpreted as the quantized compute endpoint of the frontier, not an offloading result.

Across 30 persistent-warm observations per configuration, BF16 decode throughput has mean $21.64\pm0.18$ tokens/s (sample standard deviation). The corresponding \method{} means are $15.90\pm0.51$, $17.84\pm0.57$, $21.69\pm0.56$, and $31.86\pm0.28$ tokens/s for capacities 8, 16, 32, and 64. These repeated prompt measurements characterize within-process variability; they are not independent machine replications.

\subsection{Quality relative to BF16}

\begin{table}[t]
\centering
\caption{OLMoE quality from the same checkpoint. Perplexities use the custom immutable manifests described in Section~\ref{sec:quality-eval}.}
\label{tab:quality}
\small
\begin{tabular}{lrrrr}
\toprule
Method & Wiki PPL & Dolly PPL & MC acc. norm & MC acc. raw \\
\midrule
BF16 & 17.8185 & 11.1014 & 70.8996\% & 59.0602\% \\
\method{} W4A16 & 18.5047 & 10.9399 & 70.3534\% & 58.5542\% \\
\bottomrule
\end{tabular}
\end{table}

\method{} retains 99.23\% of BF16 normalized multiple-choice accuracy:
\begin{equation}
    \frac{70.3534}{70.8996}=0.9923.
\end{equation}
This is a high relative-retention result, not an identity result. The absolute aggregate change is $-0.546$ percentage points, with a paired 95\% bootstrap interval of $[-0.964,-0.137]$ points and exact McNemar $p=0.0112$. The interval excludes zero, so the measured aggregate reduction is small but statistically detectable.

\begin{table}[t]
\centering
\caption{Paired normalized multiple-choice accuracy. Deltas are W4A16 minus BF16 in percentage points.}
\label{tab:paired}
\small
\begin{tabular}{lrrrrr}
\toprule
Task & $n$ & BF16 & W4A16 & $\Delta$ & McNemar $p$ \\
\midrule
ARC-Easy & 570 & 58.42\% & 58.77\% & $+0.35$ & 0.8877 \\
PIQA & 1,838 & 75.68\% & 75.73\% & $+0.05$ & 1.0000 \\
HellaSwag & 10,042 & 70.73\% & 70.03\% & $-0.71$ & 0.0027 \\
Aggregate & 12,450 & 70.90\% & 70.35\% & $-0.55$ & 0.0112 \\
\bottomrule
\end{tabular}
\end{table}

\begin{figure}[t]
    \centering
    \includegraphics[width=0.92\linewidth]{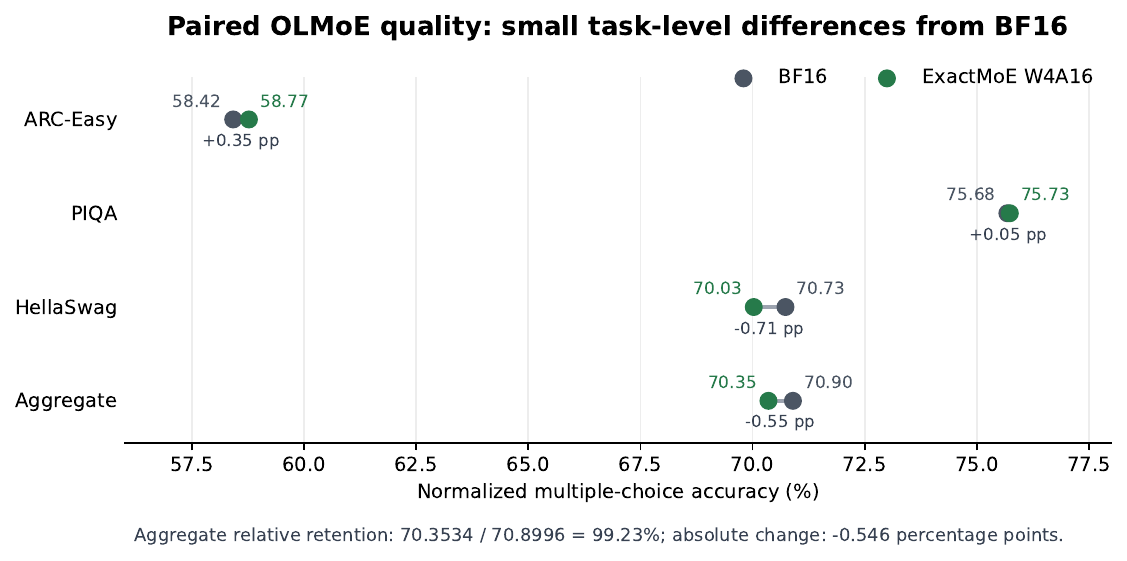}
    \caption{Paired normalized multiple-choice accuracy from the archived OLMoE evaluation. ExactMoE W4A16 is close to BF16 on all three tasks; the 12,450-example aggregate retains 99.23\% of BF16 normalized accuracy with an absolute change of $-0.546$ percentage points.}
    \label{fig:paired-quality}
\end{figure}

The aggregate change is dominated by HellaSwag because it contributes 80.7\% of the questions. ARC-Easy and PIQA show small positive point estimates with intervals crossing zero; HellaSwag shows a small negative change. WikiText perplexity increases by 3.85\%, while Dolly perplexity decreases by 1.46\%. We interpret the latter as no detectable degradation on this small custom sample, not as evidence that quantization improves the model generally. Figure~\ref{fig:paired-quality} shows the paired task-level differences.

\subsection{Fused-execution ablation}

Under a matched 16-slot cache and identical W4 weights, the retained sequential expert reference reaches 9.36 tokens/s while grouped fused execution reaches 18.43 tokens/s, a $1.97\times$ speedup. Both paths generate exactly the same 16 greedy tokens. Their last-token logits satisfy the configured tolerance, with maximum absolute difference 0.421875. The fused run records zero sequential expert calls, one sparse-layer expert invocation per decode layer, and two grouped GEMM calls per fused group.

Kernel-level tests cover gate/up and down shapes for token-group sizes from 1 to 128. The maximum relative mean error against the materialized dequantized W4 reference is 0.00560. A separate fused-MoE preflight tests global expert mapping and reports maximum absolute error 0.00098. These tests validate packing and execution; Table~\ref{tab:quality} measures the distinct W4-versus-BF16 effect.

\subsection{Batched generation}

\begin{table}[t]
\centering
\caption{Single-sweep aggregate generation throughput. \method{} uses the 16-slot configuration. This is batched generation, not continuous serving.}
\label{tab:batch}
\small
\begin{tabular}{rrrrr}
\toprule
Batch & BF16 tok/s & W4A16 tok/s & BF16 reserved GiB & W4A16 reserved GiB \\
\midrule
1 & 32.14 & 15.74 & 14.15 & 1.82 \\
2 & 46.64 & 28.26 & 14.16 & 1.83 \\
4 & 58.07 & 47.39 & 14.18 & 1.85 \\
8 & 65.25 & 90.02 & 14.68 & 1.86 \\
\bottomrule
\end{tabular}
\end{table}

Aggregate W4A16 throughput grows from 15.74 tokens/s at batch 1 to 90.02 tokens/s at batch 8. At batch 8, more routes are coalesced into grouped expert GEMMs, and W4A16 exceeds the measured BF16 aggregate throughput while using substantially less GPU memory. Because each batch point is measured once and no request scheduler is involved, these values demonstrate grouped-kernel scaling rather than production serving capacity.

\section{Discussion}

\paragraph{The useful operating point depends on the bottleneck.}
Small caches make very low-VRAM execution possible, but transfer the expert bank repeatedly. On the L4, $C=16$ moves a median 21.44~GiB per 128-token request and is 18.1\% slower than BF16 decode. Capacity 32 nearly matches BF16 throughput while retaining an 81.8\% reserved-memory reduction. Full residency is best when 4--5~GiB is available for the model; offloading is useful when even that allocation is undesirable or when a larger expert bank exceeds device capacity.

\paragraph{Kernel-native storage is central.}
The host pool is not a generic quantized checkpoint that must be transformed on every miss. It is the final MARLIN representation, including packed qweights and FP16 scales. This design minimizes miss-path software work and makes cache capacity independent of a dequantization workspace. Conversion remains an offline cost: OLMoE W4 loading and packing takes 142.7 seconds versus 4.71 seconds for BF16 loading in the measured environment.

\paragraph{``Exact'' is a systems property.}
\method{} does not preserve BF16 logits exactly. Quantized experts alter hidden states, which can alter later router decisions. The invariant is that the model's router remains intact and every expert it selects is present and executed. This differs from expert pruning, substitution, or CPU-tail schemes that change availability or execution semantics when a miss occurs.

\paragraph{High retention is not numerical identity.}
Retaining 99.23\% of baseline accuracy is a strong empirical trade-off, but the paired aggregate test detects a small reduction. The paper therefore reports retention and absolute change together. A future study should pre-register a task-weighting scheme and a non-inferiority margin rather than relying on a permissive internal pass/fail threshold.

\section{Limitations}

\method{} trades GPU memory for host capacity and data movement. Selected experts not resident on the GPU must be transferred from the packed host pool, so performance depends on routing locality, cache capacity, host-to-GPU bandwidth, and transfer latency. Limited expert reuse can cause frequent misses and eliminate offloading's throughput benefit.

The complete packed expert pool must fit in host memory. Moreover, only routed expert weights are quantized; attention, routing, embeddings, normalization, shared components, and the language-model head remain in dense precision. Memory savings therefore depend on the fraction of parameters contained in routed experts.

W4A16 does not preserve numerical identity with BF16. Quantization can alter expert outputs, hidden states, token probabilities, and later routing decisions. Although the router parameters and top-$k$ procedure remain unchanged, routing trajectories need not match BF16, and quality retention may vary across models, tasks, and experts.

The implementation requires MARLIN-compatible dimensions, grouping constraints, and NVIDIA GPUs. Other expert layouts may require padding, different packing, or another fused kernel. Supporting a new architecture also requires correctly mapping its checkpoint tensors to packed gate, up, and down projections.

Finally, \method{} targets inference with fixed weights on one device. Updating experts requires repacking them, while multi-GPU placement, distributed cache coordination, and continuous-serving schedulers require additional system support.

\section*{Reproducibility Statement}

The experimental record fixes the model and dataset revisions, random seed,
software stack, MARLIN implementation, attention backend, cache capacities,
prompt manifests, generation settings, and statistical procedures. The archive
contains raw request measurements, per-example predictions and scores, cache and
routing statistics, kernel checks, configuration files, execution logs, the
executed runtime, and SHA-256 checksums. Together, these files link the aggregate
tables and figures to the measurements from which they were derived.
Raw parity tensors are omitted to reduce archive size; parity summaries, stage
records, indices, and execution logs remain included.

\section{Ethical Considerations}

This work reduces accelerator-memory requirements for inference and can improve access to sparse language models. Lower deployment barriers can also broaden misuse of capable models. \method{} does not modify model safety behavior, training data, or licensing obligations. Deployers remain responsible for the source model's license, privacy constraints, content risks, and application-specific evaluation. The reported memory and throughput results do not establish a complete life-cycle environmental benefit.
\section{Conclusion}

\method{} combines expert-only W4A16 quantization, kernel-native host storage, configurable GPU residency, and fused grouped execution while retaining all experts. On OLMoE, cache 16 reduced peak reserved GPU memory by 87.04\% while retaining 81.85\% of BF16 decode throughput; full residency was 47.4\% faster with 71.3\% lower reserved memory. Relative normalized-accuracy retention was 99.23\%, with a small statistically detectable absolute reduction.

\clearpage
\bibliographystyle{plain}
\bibliography{references}

@article{fedus2022switch,
  title        = {Switch Transformers: Scaling to Trillion Parameter Models with Simple and Efficient Sparsity},
  author       = {Fedus, William and Zoph, Barret and Shazeer, Noam},
  journal      = {Journal of Machine Learning Research},
  volume       = {23},
  number       = {120},
  pages        = {1--39},
  year         = {2022},
  url          = {https://arxiv.org/abs/2101.03961}
}

@article{muennighoff2024olmoe,
  title        = {{OLMoE}: Open Mixture-of-Experts Language Models},
  author       = {Muennighoff, Niklas and Soldaini, Luca and Groeneveld, Dirk and Lo, Kyle and Morrison, Jacob and others},
  journal      = {arXiv preprint arXiv:2409.02060},
  year         = {2024},
  url          = {https://arxiv.org/abs/2409.02060}
}

@article{kim2023moqe,
  title        = {Mixture of Quantized Experts ({MoQE}): Complementary Effect of Low-bit Quantization and Robustness},
  author       = {Kim, Young Jin and Fahim, Raffy and Awadalla, Hany Hassan},
  journal      = {arXiv preprint arXiv:2310.02410},
  year         = {2023},
  url          = {https://arxiv.org/abs/2310.02410}
}

@article{imani2024mixture,
  title        = {Mixture of Experts with Mixture of Precisions for Tuning Quality of Service},
  author       = {Imani, HamidReza and Amirany, Abdolah and El-Ghazawi, Tarek},
  journal      = {arXiv preprint arXiv:2407.14417},
  year         = {2024},
  url          = {https://arxiv.org/abs/2407.14417}
}

@article{xue2024moeinfinity,
  title        = {{MoE-Infinity}: Efficient {MoE} Inference on Personal Machines with Sparsity-Aware Expert Cache},
  author       = {Xue, Leyang and Fu, Yao and Lu, Zhan and Sun, Chuanhao and Mai, Luo and Marina, Mahesh},
  journal      = {arXiv preprint arXiv:2401.14361},
  year         = {2024},
  url          = {https://arxiv.org/abs/2401.14361}
}

@article{tang2024hobbit,
  title        = {{HOBBIT}: A Mixed Precision Expert Offloading System for Fast {MoE} Inference},
  author       = {Tang, Peng and Liu, Jintao and Hou, Xiaolong and Pu, Yu and Wang, Jing and Heng, Pheng-Ann and Li, Chao and Guo, Minyi},
  journal      = {arXiv preprint arXiv:2411.01433},
  year         = {2024},
  url          = {https://arxiv.org/abs/2411.01433}
}

@inproceedings{kong2024swapmoe,
  title        = {{SwapMoE}: Serving Off-the-shelf {MoE}-based Large Language Models with Tunable Memory Budget},
  author       = {Kong, Rui and Li, Yuanchun and Feng, Qingtian and Wang, Weijun and Ye, Xiaozhou and Ouyang, Ye and Kong, Linghe and Liu, Yunxin},
  booktitle    = {Proceedings of the 62nd Annual Meeting of the Association for Computational Linguistics (Volume 1: Long Papers)},
  pages        = {6710--6720},
  year         = {2024},
  doi          = {10.18653/v1/2024.acl-long.363},
  url          = {https://aclanthology.org/2024.acl-long.363/}
}

@inproceedings{kamahori2025fiddler,
  title        = {Fiddler: {CPU--GPU} Orchestration for Fast Inference of Mixture-of-Experts Models},
  author       = {Kamahori, Keisuke and Tang, Tian and Gu, Yile and Zhu, Kan and Kasikci, Baris},
  booktitle    = {International Conference on Learning Representations},
  year         = {2025},
  url          = {https://arxiv.org/abs/2402.07033}
}

@inproceedings{cao2025moelightning,
  title        = {{MoE-Lightning}: High-Throughput {MoE} Inference on Memory-constrained {GPUs}},
  author       = {Cao, Shiyi and Liu, Shu and Griggs, Tyler and Schafhalter, Peter and Liu, Xiaoxuan and Sheng, Ying and Gonzalez, Joseph E. and Zaharia, Matei and Stoica, Ion},
  booktitle    = {Proceedings of the 30th ACM International Conference on Architectural Support for Programming Languages and Operating Systems},
  year         = {2025},
  doi          = {10.1145/3669940.3707267},
  url          = {https://arxiv.org/abs/2411.11217}
}

@inproceedings{duanmu2025mxmoe,
  title        = {{MxMoE}: Mixed-precision Quantization for {MoE} with Accuracy and Performance Co-Design},
  author       = {Duanmu, Haojie and Li, Xiuhong and Yuan, Zhihang and Zheng, Size and Duan, Jiangfei and Zhang, Xingcheng and Lin, Dahua},
  booktitle    = {Proceedings of the 42nd International Conference on Machine Learning},
  series       = {Proceedings of Machine Learning Research},
  volume       = {267},
  pages        = {14793--14806},
  year         = {2025},
  url          = {https://arxiv.org/abs/2505.05799}
}

@article{huang2025milo,
  title        = {{MiLo}: Efficient Quantized {MoE} Inference with Mixture of Low-Rank Compensators},
  author       = {Huang, Beichen and Yuan, Yueming and Shao, Zelei and Zhang, Minjia},
  journal      = {arXiv preprint arXiv:2504.02658},
  year         = {2025},
  url          = {https://arxiv.org/abs/2504.02658}
}

@article{chu2025dynaexq,
  title        = {Dynamic Expert Quantization for Scalable Mixture-of-Experts Inference},
  author       = {Chu, Kexin and Xiang, Dawei and Shen, Zixu and Yang, Yiwei and Liu, Zecheng and Zhang, Wei},
  journal      = {arXiv preprint arXiv:2511.15015},
  year         = {2025},
  url          = {https://arxiv.org/abs/2511.15015}
}

@article{yang2026pagedweight,
  title        = {{PagedWeight}: Efficient {MoE} {LLM} Serving with Dynamic Quality-Aware Weight Quantization},
  author       = {Yang, Yuchen and Zhao, Yifan and Dasgupta, Anisha and Misailovic, Sasa},
  journal      = {arXiv preprint arXiv:2607.16184},
  year         = {2026},
  url          = {https://arxiv.org/abs/2607.16184}
}

@article{frantar2024marlin,
  title        = {{MARLIN}: Mixed-Precision Auto-Regressive Parallel Inference on Large Language Models},
  author       = {Frantar, Elias and Castro, Roberto L. and Chen, Jiale and Hoefler, Torsten and Alistarh, Dan},
  journal      = {arXiv preprint arXiv:2408.11743},
  year         = {2024},
  url          = {https://arxiv.org/abs/2408.11743}
}

@inproceedings{kwon2023vllm,
  title        = {Efficient Memory Management for Large Language Model Serving with {PagedAttention}},
  author       = {Kwon, Woosuk and Li, Zhuohan and Zhuang, Siyuan and Sheng, Ying and Zheng, Lianmin and Yu, Cody Hao and Gonzalez, Joseph E. and Zhang, Hao and Stoica, Ion},
  booktitle    = {Proceedings of the 29th Symposium on Operating Systems Principles},
  pages        = {611--626},
  year         = {2023},
  url          = {https://arxiv.org/abs/2309.06180}
}

@article{frantar2022gptq,
  title        = {{GPTQ}: Accurate Post-Training Quantization for Generative Pre-trained Transformers},
  author       = {Frantar, Elias and Ashkboos, Saleh and Hoefler, Torsten and Alistarh, Dan},
  journal      = {arXiv preprint arXiv:2210.17323},
  year         = {2022},
  url          = {https://arxiv.org/abs/2210.17323}
}

@inproceedings{lin2024awq,
  title        = {{AWQ}: Activation-aware Weight Quantization for {LLM} Compression and Acceleration},
  author       = {Lin, Ji and Tang, Jiaming and Tang, Haotian and Yang, Shang and Dang, Xingyu and Han, Song},
  booktitle    = {Proceedings of Machine Learning and Systems},
  year         = {2024},
  url          = {https://arxiv.org/abs/2306.00978}
}

@article{bai2023qwen,
  title        = {Qwen Technical Report},
  author       = {Bai, Jinze and Bai, Shuai and Chu, Yunfei and Cui, Zeyu and Dang, Kai and others},
  journal      = {arXiv preprint arXiv:2309.16609},
  year         = {2023},
  url          = {https://arxiv.org/abs/2309.16609}
}

@misc{qwen2024moe,
  title        = {{Qwen1.5-MoE}: Matching 7B Model Performance with $1/3$ Activated Parameters},
  author       = {{Qwen Team}},
  year         = {2024},
  howpublished = {Technical blog},
  url          = {https://qwenlm.github.io/blog/qwen-moe/}
}

@misc{qwen2026qwen36,
  title        = {{Qwen3.6-35B-A3B}: Agentic Coding Power, Now Open to All},
  author       = {{Qwen Team}},
  year         = {2026},
  howpublished = {Technical blog},
  url          = {https://qwen.ai/blog?id=qwen3.6-35b-a3b}
}

@article{clark2018arc,
  title        = {Think You Have Solved Question Answering? Try {ARC}, the {AI2} Reasoning Challenge},
  author       = {Clark, Peter and Cowhey, Isaac and Etzioni, Oren and Khot, Tushar and Sabharwal, Ashish and Schoenick, Carissa and Tafjord, Oyvind},
  journal      = {arXiv preprint arXiv:1803.05457},
  year         = {2018},
  url          = {https://arxiv.org/abs/1803.05457}
}

@inproceedings{bisk2020piqa,
  title        = {{PIQA}: Reasoning about Physical Commonsense in Natural Language},
  author       = {Bisk, Yonatan and Zellers, Rowan and Le Bras, Ronan and Gao, Jianfeng and Choi, Yejin},
  booktitle    = {Proceedings of the AAAI Conference on Artificial Intelligence},
  year         = {2020},
  url          = {https://arxiv.org/abs/1911.11641}
}

@inproceedings{zellers2019hellaswag,
  title        = {{HellaSwag}: Can a Machine Really Finish Your Sentence?},
  author       = {Zellers, Rowan and Holtzman, Ari and Bisk, Yonatan and Farhadi, Ali and Choi, Yejin},
  booktitle    = {Proceedings of the 57th Annual Meeting of the Association for Computational Linguistics},
  year         = {2019},
  url          = {https://arxiv.org/abs/1905.07830}
}

@article{merity2016wikitext,
  title        = {Pointer Sentinel Mixture Models},
  author       = {Merity, Stephen and Xiong, Caiming and Bradbury, James and Socher, Richard},
  journal      = {arXiv preprint arXiv:1609.07843},
  year         = {2016},
  url          = {https://arxiv.org/abs/1609.07843}
}

@misc{conover2023dolly,
  title        = {Free Dolly: Introducing the World's First Truly Open Instruction-Tuned {LLM}},
  author       = {Conover, Mike and Hayes, Matt and Mathur, Ankit and Meng, Xiangrui and Xie, Jianwei and Wan, Jun and Shah, Sam and Ghodsi, Ali and Wendell, Patrick and Zaharia, Matei and Xin, Reynold},
  year         = {2023},
  howpublished = {Databricks technical report and dataset release},
  url          = {https://www.databricks.com/blog/2023/04/12/dolly-first-open-commercially-viable-instruction-tuned-llm}
}
\end{document}